\documentclass[11pt]{article}
\usepackage[latin9]{inputenc}
\usepackage{booktabs}

\providecommand{\tabularnewline}{\\}

\usepackage{coling2016}
\usepackage{times}
\usepackage{url}
\usepackage{latexsym}
\usepackage{multirow}
\usepackage{amssymb}
\usepackage{wasysym}
\usepackage{float}
\usepackage{graphicx}
\usepackage{bm}
\title{Neural Paraphrase Generation with Stacked Residual LSTM Networks}

\author{Aaditya Prakash$^{1,2}$, Sadid A. Hasan$^2$, Kathy Lee$^2$, Vivek Datla$^2$, \\\bf{Ashequl Qadir$^2$, Joey Liu$^2$, Oladimeji Farri$^2$} \\
$^1$Brandeis University, Waltham, MA, USA \\
$^2$Artificial Intelligence Laboratory, Philips Research North America, Cambridge, MA, USA \\
{\tt \{aprakash,aaditya.prakash\}@\{brandeis.edu,philips.com\}} \\
{\tt \{sadid.hasan,kathy.lee\_1,vivek.datla\}@philips.com} \\
{\tt \{ashequl.qadir,joey.liu,dimeji.farri\}@philips.com}}

\begin{document}
\maketitle 
					
\begin{abstract}
In this paper, we propose a novel neural approach for paraphrase generation. Conventional paraphrase generation methods either leverage hand-written rules and thesauri-based alignments, or use statistical machine learning principles. To the best of our knowledge, this work is the first to explore deep learning models for paraphrase generation. Our primary contribution is a stacked residual LSTM network, where we add residual connections between LSTM layers. This allows for efficient training of deep LSTMs. We evaluate our model and other state-of-the-art deep learning models on three different datasets: \texttt{PPDB}, \texttt{WikiAnswers}, and \texttt{MSCOCO}. Evaluation results demonstrate that our model outperforms sequence to sequence, attention-based, and bi-directional LSTM models on \texttt{BLEU}, \texttt{METEOR}, \texttt{TER}, and an \emph{embedding}-based sentence similarity metric.

\end{abstract}

\section{Introduction}

\blfootnote{\hspace{-0.65cm} This work is licensed under a Creative Commons Attribution 4.0 International License. License details: \\\url{http://creativecommons.org/licenses/by/4.0/}}

Paraphrasing, the act to express the same meaning in different possible ways, is an important subtask in various Natural Language Processing (NLP) applications such as question answering, information extraction, information retrieval, summarization and natural language generation. Research on paraphrasing methods typically aims at solving three related problems: (1) recognition (i.e.\ to identify if two textual units are paraphrases of each other), (2) extraction (i.e.\ to extract paraphrase instances from a thesaurus or a corpus), and (3) generation (i.e.\ to generate a reference paraphrase given a source text) \cite{Madnani2010}. In this paper, we focus on the paraphrase generation problem. 

Paraphrase generation has been used to gain performance improvements in several NLP applications, for example, by generating query variants or pattern alternatives for information retrieval, information extraction or question answering systems, by creating reference paraphrases for automatic evaluation of machine translation and document summarization systems, and by generating concise or simplified information for sentence compression or sentence simplification systems \cite{Madnani2010}. Traditional paraphrase generation methods exploit hand-crafted rules \cite{McKeown1983} or automatically learned complex paraphrase patterns \cite{Zhao2009}, use thesaurus-based \cite{Hassan2007} or semantic analysis driven natural language generation approaches \cite{Kozlowski2003}, or leverage statistical machine learning theory \cite{quirk2004,Wubben2010}. In this paper, we propose to use deep learning principles to address the paraphrase generation problem.
 
Recently, techniques like sequence to sequence learning \cite{SutskeverVL14} have been applied to various NLP tasks with promising results, for example, in the areas of machine translation \cite{cho14,Bahdanau15}, speech recognition \cite{li2015constructing}, language modeling \cite{vinyals2015grammar}, and dialogue systems \cite{serban2016multiresolution}. Although paraphrase generation can be formulated as a sequence to sequence learning task, not much work has been done in this area with regard to applications of state-of-the-art deep neural networks. There are several works on paraphrase recognition \cite{SocherEtAl2011,yin-schutze2015,kiros2015skip}, but those employ classification techniques and do not attempt to generate paraphrases. More recently, attention-based Long Short-Term Memory (LSTM) networks have been used for textual entailment generation \cite{KolesnykRR16}; however, paraphrase generation is a type of bi-directional textual entailment generation and no prior work has proposed a deep learning-based formulation of this task.
 
To address this gap in the literature, we explore various types of sequence to sequence models for paraphrase generation. We test these models on three different datasets and evaluate them using well recognized metrics. Along with the application of various existing sequence to sequence models for the paraphrase generation task, in this paper we also propose a new model that allows for training multiple stacked LSTM networks by introducing a residual connection between the layers. This is inspired by the recent success of such connections in a deep Convolutional Neural Network (CNN) for the image recognition task \cite{he2015deep}. Our experiments demonstrate that the proposed model can outperform other techniques we have explored.
 
Most of the deep learning models for NLP use Recurrent Neural Networks (RNNs). RNNs differ from normal perceptrons as they allow gradient propagation in time to model sequential data with variable-length input and output \cite{SutskeverMH11}. In practice, RNNs often suffer from the vanishing/exploding gradient problems while learning long-range dependencies \cite{Bengio94}. LSTM \cite{Hochreiter:1997} and GRU \cite{cho14} are known to be successful remedies to these problems.
 
It has been observed that increasing the depth of a deep neural network can improve the performance of the model \cite{simonyan2014very,he2015deep} as deeper networks learn better representations of features \cite{farabet2013learning}. In the vision-related tasks where CNNs are more widely used, adding many layers of neurons is a common practice. For tasks like speech recognition \cite{li2015constructing} and also in machine translation, it is useful to stack layers of LSTM or other variants of RNN.\ So far this has been limited to only a few layers due to the difficulty in training deep RNN networks. We propose to add residual connections between multiple stacked LSTM networks and show that this allows us to stack more layers of LSTM successfully.

The rest of the paper is organized as follows: Section 2 presents a brief overview of the sequence to sequence models followed by a description of our proposed residual deep LSTM model, Section 3 describes the datasets used in this work, Section 4 explains the experimental setup, Section 5 presents the evaluation results and analyses, Section 6 discusses the related work, and in Section 7 we conclude and discuss future work.

\section{Model Description}
\subsection{Encoder-Decoder Model}
\begin{figure}
    \centering
    \includegraphics[scale=0.5]{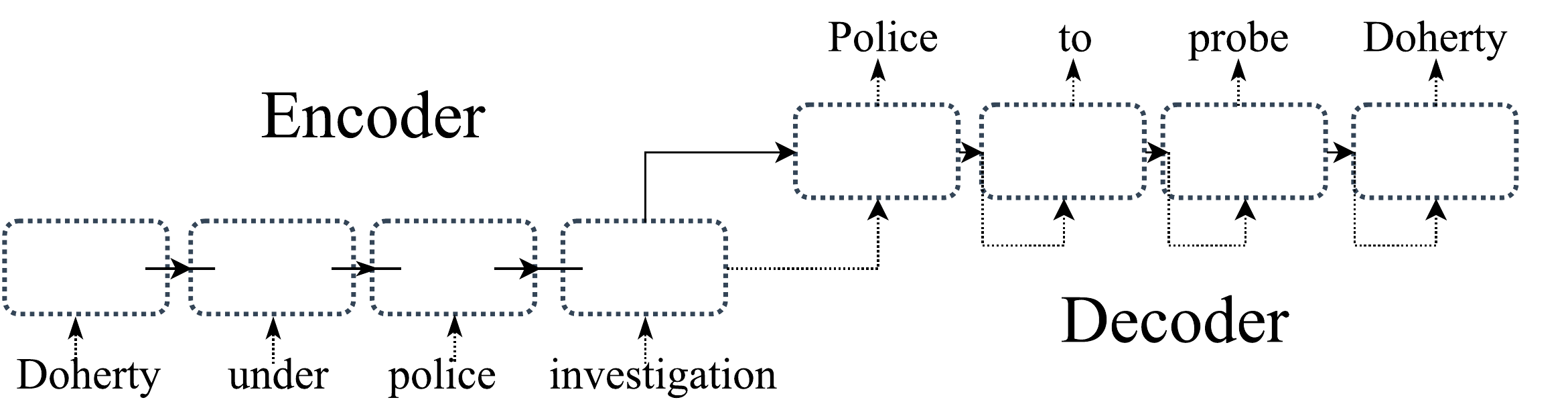}
     \caption{\label{fig:enc_dec} Encoder-Decoder framework for sequence to sequence learning.}
		\vspace{3mm}
\end{figure}

A neural approach to sequence to sequence modeling proposed by Sutskever et al.~\shortcite{SutskeverVL14} is a two-component model, where a source sequence is first encoded into some low dimensional representation (Figure~\ref{fig:enc_dec}) that is later used to reproduce the sequence back to a high dimensional target sequence (i.e.\ decoding). In machine translation, an encoder operates on a sentence written in the source language and encodes its meaning to a vector representation before the decoder can take that vector (which represents the meaning) and generate a sentence in the target language. These encoder-decoder blocks can be either a vanilla RNN or its variants. While producing the target sequence, the generation of each new word depends on the model and the preceding generated word. Generation of the first word in the target sequence depends on the special `\texttt{EOS}' (end-of-sentence) token appended to the source sequence.
 
The training objective is to maximize the log probability of the target sequence given the source sequence. Therefore, the best possible decoded target is the one that has the maximum score over the length of the sequence. To find this, a small set of hypotheses (candidate set) called \emph{beam size} is used and the total score for all these hypotheses are computed. In the original work by Sutskever et al.~\shortcite{SutskeverVL14}, they observe that although a beam size of 1 achieves good results, a higher beam size is always better. This is because for some of the hypotheses, the first word may not always have the highest score.
 
\subsection{Deep LSTM}
\begin{figure} 
\begin{minipage}{0.45\textwidth}
\begin{enumerate}
  \item
  \phantom \\
  Gates
 
        \(\displaystyle   i_{t} = \sigma(W_{xi}x_{t} + W_{hi}h_{t-1} + b_{i}) \) \\
        \(\displaystyle   f_{t} = \sigma(W_{xf}x_{t} + W_{hf}h_{t-1} + b_{f}) \) \\
        \(\displaystyle   o_{t} = \sigma(W_{xo}x_{t} + W_{ho}h_{t-1} + b_{o}) \) \\
  \item Input transform
 
        \(\displaystyle    c\_in_{t} = \textrm{tanh}(W_{xc}x_{t} + W_{hc}h_{t-1} + b_{c\_in})  \) \\
 
  \item State Update
 
        \(\displaystyle    c_{t} = f_{t} \ \astrosun \ c_{t-1} + i_{t} \ \astrosun c\_in_{t} \) \\
        \(\displaystyle    h_{t} = o_{t} \ \astrosun \ \textrm{tanh}(c_{t}) \) \\

\end{enumerate}
\end{minipage}%
\hfill
\begin{minipage}{0.45\textwidth}
\begin{tabular}{p{\textwidth}}
    \includegraphics[scale=0.4]{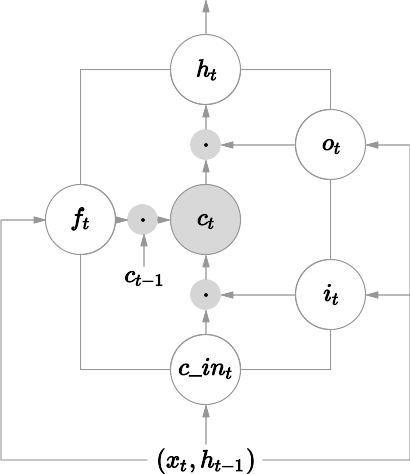}
		\caption{\label{fig:lstmcell} LSTM cell \cite{Paszke15}.}
		\vspace{3mm}
		\end{tabular}
\end{minipage}%
\end{figure}

LSTM (Figure~\ref{fig:lstmcell}) is a variant of RNN, which computes the hidden state $h_t$ using a different approach by adding an internal memory cell $c_t \! \in \! \mathbb{R}^n$ at every time step $t$. In particular, an LSTM unit considers the input state $x_t$ at time step $t$, the hidden state $h_{t-1}$, and the internal memory state $c_{t-1}$ at time step $t-1$ to produce the hidden state $h_t$ and the internal memory state $c_t$ at time step $t$. The memory cell is controlled via three learned gates: input $i$, forget $f$, and output $o$. These memory cells use the addition of gradient with respect to time and thus minimize the gradient explosion. In most NLP tasks, LSTM outperforms vanilla RNN \cite{sundermeyer2012lstm}. Therefore, for our model we only explore LSTM as a basic unit in the encoder and decoder. Here, we describe the basic computations in an LSTM unit, which will provide the grounding to understand the residual connections between stacked LSTM layers later.
 
In the equations above, $W_{x\_}, W_{h\_}$ are the learned parameters for $x$ and $h$ respectively. $\sigma(.)$ and $\tanh(.)$ denote element-wise sigmoid and hyperbolic tangent functions respectively. $\astrosun$ is the element-wise multiplication operator and $b$ denotes the added bias.

Graves~\shortcite{graves2013generating} explored the advantages of deep LSTMs for handwriting recognition and text generation. There are multiple ways of combining one layer of LSTM with another. For example, Pascanu et al.~\shortcite{pascanu2013construct} explored multiple ways of combining them and discussed various difficulties in training deep LSTMs. In this work, we employ vertical stacking where only the output of the previous layer of LSTM is fed to the input, as compared to
the stacking technique used by Sutskever et al.~\shortcite{SutskeverVL14}, where hidden states of all LSTM layers are fully connected. In our model, all but the first layer input at time step $t$ is passed from the hidden state of the previous layer $h_{t}^{l}$, where $l$ denotes the layer. This is similar to stacked RNN proposed by Bengio et al.~\shortcite{Bengio94} but with LSTM units. Thus, for a layer $l$ the activation is described by:
 
$$
\bm{h}_{t}^{(l)} = f_{h}^{l}(\bm{h}_{t}^{(l-1)}, \bm{h}_{t-1}^{(l)})
$$
 
where hidden states $\bm{h}$ are recursively computed and $\bm{h}_{t}^{(l)} $ at $t=0$ and $l=0$ is given by the LSTM equation of $h_{t}$.

\subsection{Stacked Residual LSTM}
We take inspiration from a very successful deep learning network \emph{ResNet} \cite{he2015deep} with regard to adding residue for the purpose of learning. With  theoretical and empirical reasoning, He et al.~\shortcite{he2015deep} have shown that the explicit addition of the residue $x$ to the function being learned allows
for deeper network training without overfitting the data. 
 
\begin{figure}
    \centering
    \includegraphics[scale=0.5]{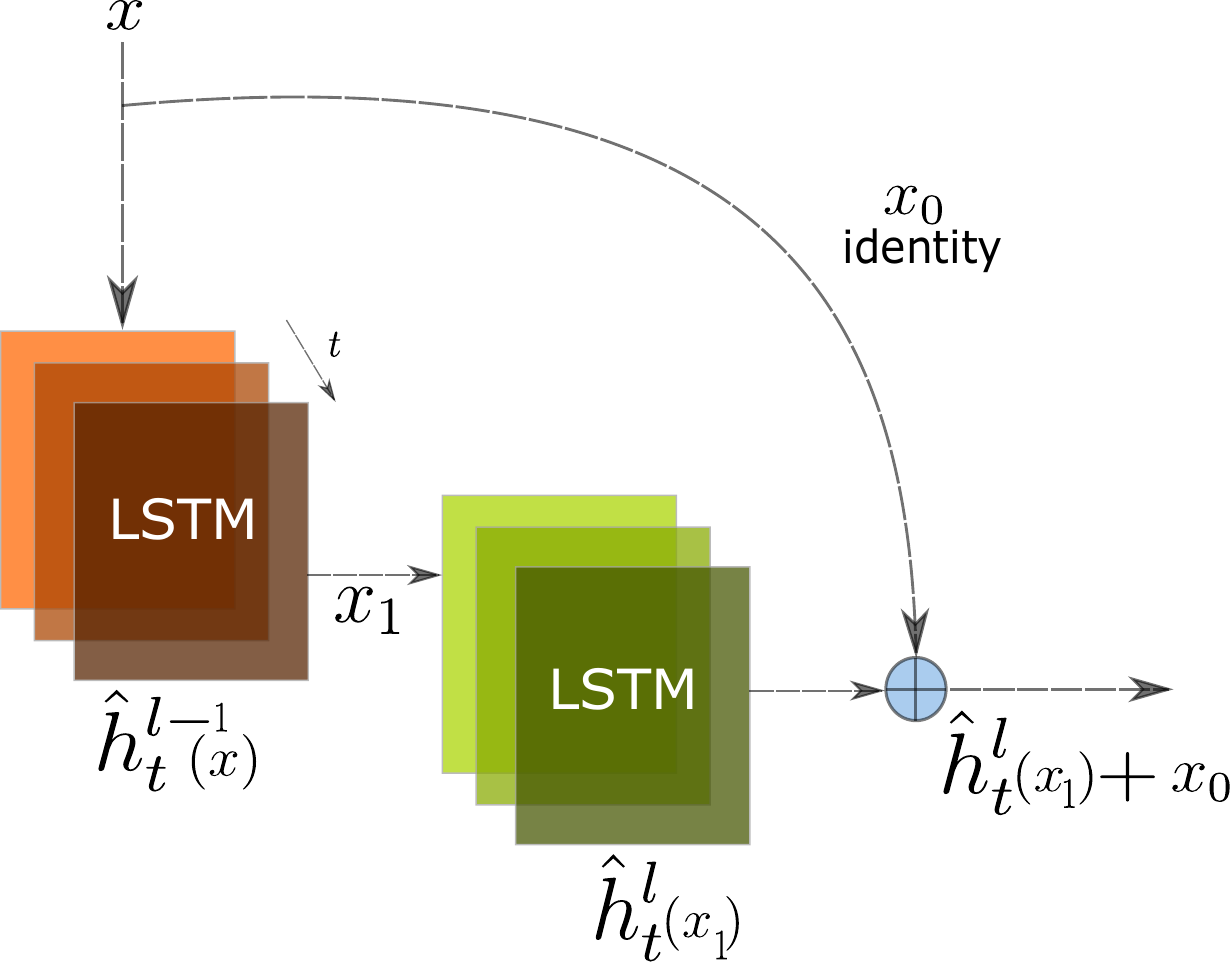}
      \caption{\label{fig:res} A unit of stacked residual LSTM.}
		\vspace{3mm}
\end{figure}

When stacking multiple layers of neurons, the network often suffers through a \emph{degradation} problem \cite{he2015deep}. The degradation problem arises due to the low convergence rate of training error and is different from the vanishing gradient problem. Residual connections can help overcome this issue. We experimented with four-layers of stacked LSTM for each of the model. Residual connections are added at layer two as the pointwise addition (see Figure~\ref{fig:res}), and thus it requires the input to be in the same dimension as the output of $h_t$. Principally because of this reason, we use a simple last hidden unit stacking of LSTM instead of a more intricate way as shown by Sutskever et al.~\shortcite{SutskeverVL14}. This allowed us to clip the $h_t$ to match the dimension of $x_{t-2}$ where they were not the same. Similar results could be achieved by padding $x$ to match the dimension instead. The function $\bm{\hat{h}}$ that is being learned for the layer with residual connection is therefore:
 
$$
\bm{\hat{h}}_{t}^{(l)} = f_{h}^{l}(\bm{h}_{t}^{(l-1)}, \bm{h}_{t-1}^{(l)}) + x_{l-n}
$$
where $\hat{h}$ for layer $l$ is updated with residual value $x_{l-n} $ and $x_{i}$ represents the input to layer $i+1$. Residual connection is added after every $n$ layers. However, for stacked LSTM, $n>3$ is very expensive in terms of computation. In this paper we experimented with $n=2$. Note that, when $n=1$, the resulting function learned is a standard LSTM with bias that depends on the input $x$. That is why, it is not necessary to add the residual connection after every stacked layer of LSTM.\ The addition of residual connection does not add any learnable parameters. Therefore, this does not increase the complexity of the model unlike bi-directional models which double the number of LSTM units.
 
\section{Datasets}
We present the performance of our model on three datasets, which are significantly different in their characteristics. So, evaluating our paraphrase generation approach on these datasets demonstrates the versatility and robustness of our model.
 
\textbf{PPDB} \cite{pavlick2015} is a well known paraphrase dataset used for various NLP tasks. It comes in different sizes and the precision of the paraphrases degrades with the size of the dataset. We use the size $L$ dataset from \texttt{PPDB} 2.0, which comes with over $18M$ paraphrases including lexical, phrasal and syntactic types. We have omitted the syntactic paraphrases and the instances which contain numbers, as they increase the vocabulary size significantly without giving any advantage of a larger dataset. This dataset contains relatively short paraphrases ($86\%$ of the data is less than four words), which makes it suitable for synonym generation and phrase substitution to address lexical and phrasal paraphrasing \cite{Madnani2010}. For some phrases, \texttt{PPDB} has one-to-many paraphrases. We collect all such phrases to make a set of paraphrases and sampling without replacement was used to obtain the source and reference phrases. 

 
\textbf{WikiAnswers} \cite{fader2013paraphrase} is a large question paraphrase corpus created by crawling the WikiAnswers website\footnote{http://wiki.answers.com}, where users can post questions and answers about any topic. The paraphrases are different questions, which were tagged by the users as similar questions. The dataset contains approximately 18M word-aligned question pairs. Sometimes, there occurs a loss of specialization between a given source question and its corresponding reference question when a paraphrase is tagged as similar to a reference question. For example, ``prepare a \emph{three month} cash budget'' is tagged to ``how to prepare a cash budget''. This happens because general questions are typically more popular and get answered. So, specific questions are redirected to the general ones due to a comparative lack of interest in the very specific questions. It should be noted that this dataset comes preprocessed and lemmatized. We refer the reader to the original paper for more details.
 
\textbf{MSCOCO} \cite{lin2014microsoft} dataset contains human annotated captions of over $120K$ images. Each image contains five captions from five different annotators. While there is no guarantee that the human annotations are paraphrases, the nature of the images (which tends to focus on only a few objects and in most cases one prominent object or action) allows most annotators describe the most obvious things in an image. In fact, this is the main reason why neural networks for generating captions obtain better \texttt{BLEU} scores \cite{VinyalsTBE14}, which confirms the suitability of using this dataset for the paraphrase generation task.

 
\section{Experimental Settings}
\subsection{Data Selection}
For \texttt{PPDB} we remove the phrases that contain numbers including all syntactic phrases. This gives us a total of $5.3M$ paraphrases from which we randomly select $90\%$ instances for training. For testing, we randomly select $20K$ pairs of paraphrases from the remaining $10\%$ data. Although \texttt{WikiAnswers} comes with over $29M$ instances, we randomly select $4.8M$ for training to keep the training size similar to \texttt{PPDB} (see Table 1). $20K$ instances were randomly selected from the remaining data for testing. Note that, for the \texttt{WikiAnswers} dataset, we clip the vocabulary size\footnote{\texttt{WikiAnswers} dataset had many spelling errors yielding a very large vocabulary size (approximately $250K$). Hence, we selected the most frequent $50K$ words in the vocabulary to reduce the computational complexity.} to $50K$ and use the special \emph{UNK} symbol for the words outside the vocabulary. \texttt{MSCOCO} dataset has five captions for every image. This dataset comes with separate subsets for training and validation: \emph{Train 2014} contains over $82K$ images and \emph{Val 2014} contains over $40K$ images. From the five captions accompanying each image, we randomly omit one caption and use the other four as training instances (by creating two source-reference pairs). Thus, we obtain a collection of over $330K$ instances for training and $20K$ instances for testing. Because of the free form nature of the caption generation task \cite{VinyalsTBE14}, some captions were very long. We reduced those captions to the size of $15$ words (by removing the words beyond the first $15$) in order to reduce the training complexity of the models.
 
\begin{table}[!htb]
    \begin{minipage}{.5\linewidth}
      \centering
 
    \small
    \begin{tabular}{lllr}
    \midrule
    \multicolumn{1}{c}{\textbf{Dataset}} & \multicolumn{1}{c}{\textbf{Training}} & \multicolumn{1}{c}{\textbf{Test}} & \multicolumn{1}{c}{\textbf{Vocabulary Size}}\\
    \midrule
    \texttt{PPDB}                        & 4,826,492                    & 20,000         & 38,279          \\
    \texttt{WikiAnswers}                 & 4,826,492                    & 20,000         & 50,000          \\
    \texttt{MSCOCO}                      & 331,163                      & 20,000         & 30,332         \\
    \midrule
    \end{tabular}
     \caption{Dataset details.}
 
    \end{minipage}%
    \begin{minipage}{.5\linewidth}
      \centering
 
        \small
        \begin{tabular}{ll}
        \midrule
        \textbf{Models}                                      & \textbf{Reference}  \\
        \midrule
        Sequence to Sequence                                 & \cite{SutskeverVL14} \\
        With Attention                                       & \cite{Bahdanau15} \\
        Bi-directional LSTM                                  & \cite{graves2013hybrid} \\
        Residual LSTM                                        & Our proposed model                                   \\              
        \midrule
        \end{tabular}
        \caption{Models.}
    \end{minipage}
\end{table}

\subsection{Models}
We experimented with four different models (see Table 2). For each model, we experimented with two- and four-layers of stacked LSTMs. This was motivated by the state-of-the-art speech recognition systems that also use three to four layers of stacked LSTMs \cite{li2015constructing}. In encoder-decoder models, the size of the beam search used during inference is very important. Larger beam size always gives higher accuracy but is associated with a computational cost. We experimented with beam sizes of $5$ and $10$ to compare the models, as these are the most common beam sizes used in the literature \cite{SutskeverVL14}. The bi-directional model used half of the number of layers shown for other models. This was done to ensure similar parameter sizes across the models.

\subsection{Training}
We used a one-hot vector approach to represent the words in all models. Models were trained with a stochastic gradient descent (SGD) algorithm. The learning rate began at $1.0$, and was halved after every third training epoch. Each network was trained for ten epochs. In order to allow exploration of a wide variety of models, training was restricted to a limited number of epochs, and no hyper-parameter search was performed. A standard dropout \cite{srivastava2014dropout} of 50\% was applied after every LSTM layer. The number of LSTM units in each layer was fixed to $512$ across all models. Training time ranged from $36$ hours for \texttt{WikiAnswers} and \texttt{PPDB} to $14$ hours for \texttt{MSCOCO} on a \emph{Titan X} with \emph{CuDNN 5} using Theano version $0.9.0dev1$ \cite{2016arXiv160502688short}. 

\begin{figure}
    \centering
    \includegraphics[scale=0.20]{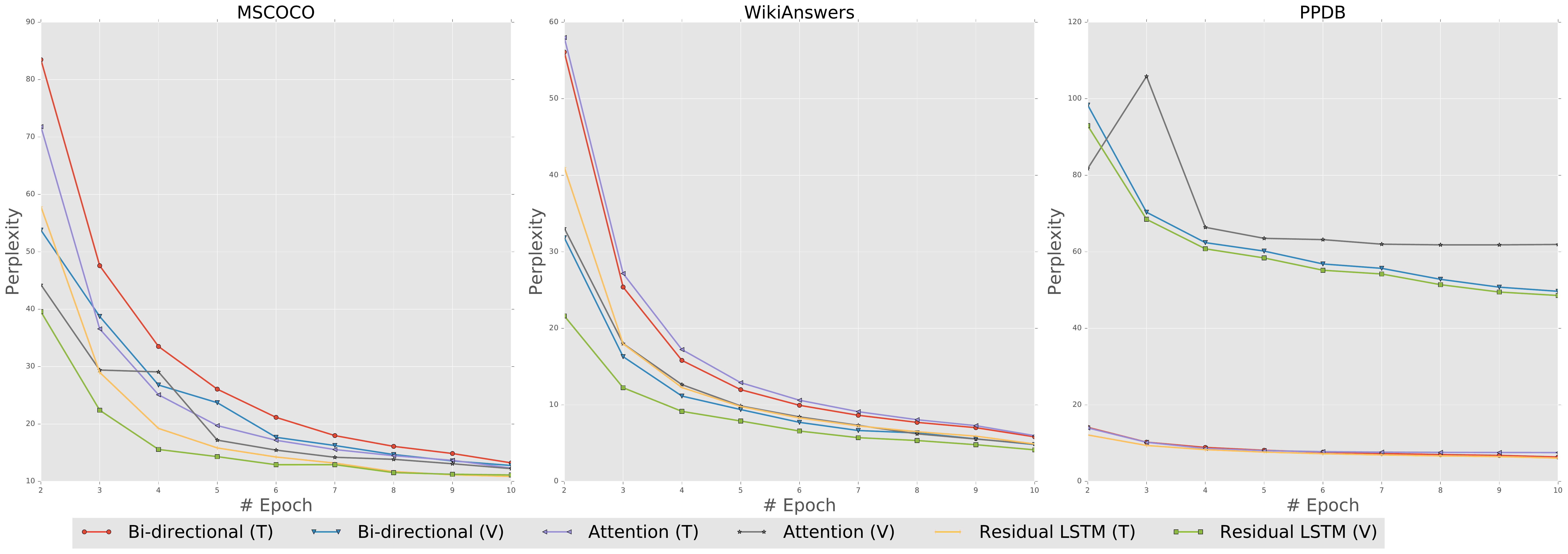}
       \caption{\label{fig:perplex} Perplexity during training ($T$) and validation ($V$) for various models [shared legend]. A lower perplexity represents a better model.}
		\vspace{3mm}
\end{figure}

A beam search algorithm was used to generate optimal paraphrases by exploiting the trained models in the testing phase \cite{SutskeverVL14}. We used perplexity as the loss function during training. Perplexity measures the uncertainty of the language model, corresponding to how many bits on average would be needed to encode each word given the language model. A lower perplexity indicates a better score. While \texttt{WikiAnswers} and \texttt{MSCOCO} had a very good correlation between training and validation perplexity, overfitting was observed with \texttt{PPDB} that yielded a worse validation perplexity (see Figure~\ref{fig:perplex}).
 
\section{Evaluation}
\subsection{Metrics}
To quantitatively evaluate the performance of our paraphrase generation models, we use the well-known automatic evaluation metrics\footnote{We used the software available at \url{https://github.com/jhclark/multeval}} for comparing parallel corpora:
\texttt{BLEU} \cite{papineni2002}, \texttt{METEOR}  \cite{Lavie2007}, and Translation Error Rate (\texttt{TER}) \cite{snover2006study}. Even though these metrics were designed for machine translation, previous works have shown that they can perform well for the paraphrase recognition task \cite{Madnani2012} and correlate well with human judgments in evaluating generated paraphrases \cite{Wubben2010}. 

Although there exists a few automatic evaluation metrics that are specifically designed for paraphrase generation, such as \texttt{PEM} (Paraphrase Evaluation Metric) \cite{LiuDN10} and \texttt{PINC} (Paraphrase In N-gram Changes) \cite{ChenD11}, they have certain limitations. \texttt{PEM} relies on large in-domain bilingual parallel corpora along with sample human ratings for training while it can only model paraphrasing up to the phrase-level granularity. \texttt{PINC} attempts to solve these limitations by proposing a method that is essentially the inverse of \texttt{BLEU}, as it calculates the n-gram difference between the source and the reference sentences. Although \texttt{PINC} correlates well with human judgments in lexical dissimilarity assessment, \texttt{BLEU} has been shown to correlate better for semantic equivalence agreements at the sentence-level when a sufficiently large number of reference sentences are available for each source sentence \cite{ChenD11}.

\texttt{BLEU} considers exact matching between reference paraphrases and system generated paraphrases by considering n-gram overlaps while \texttt{METEOR} improves upon this measure via stemming and synonymy using WordNet. \texttt{TER} measures the number of edits required to change a system generated paraphrase into one of the reference paraphrases. As suggested in Clark et al.~\shortcite{Clark:2011}, we used a stratified approximate randomization (AR) test. AR calculates the probability of a metric score providing the same reference sentence by chance. We report our {\it p}-values at \texttt{95\%} Confidence Intervals (CI).
 
The major limitation of these evaluation metrics is that they do not consider the meaning of the paraphrases, and hence, are not able to capture paraphrases of entities. For example, these metrics do not reward the paraphrasing of \emph{``London''} to \emph{``Capital of UK''}. Therefore, we also evaluate our models on a sentence similarity metric\footnote{We used the software available at \url{https://github.com/julianser/hed-dlg-truncated/}} proposed by Rus et al.~\shortcite{rus2012comparison}. This metric uses word embeddings to compare the phrases. In our experiments, we used Word2Vec embeddings pre-trained on the Google News Corpus \cite{mikolov2014word2vec}. This is referred to as \emph{`Emb Greedy'} in our results table.
 
\subsection{Results}
Table 3 presents the results from various models across different datasets. $\uparrow$ denotes that higher scores represent better models while $\downarrow$ means that a lower score yields a better model. Although our focus is on stacked residual LSTM, which is applicable only when there are more than two layers, we still present the scores from two-layer LSTM as a baseline. This provides a good comparison against deeper models. The results demonstrate that our proposed model outperforms other models on \texttt{BLEU}  and \texttt{TER} for all datasets. On \emph{Emb Greedy}, our model outperforms other models in all datasets except the Attention model when beam size is 10. On \texttt{METEOR}, our model outperforms other models on \texttt{MSCOCO} and \texttt{WikiAnswers}; however, for \texttt{PPDB}, the simple sequence to sequence model performs better. Note that these results were obtained by using single models and no ensemble of the models was used. 

To calculate \texttt{BLEU} and \texttt{METEOR}, four references were used for \texttt{MSCOCO}, and five for \texttt{PPDB} and \texttt{WikiAnswers}. In some instances, \texttt{WikiAnswers} did not have up to five reference paraphrases for every source, hence, those were calculated on reduced references. In Table 4, we present the variance due to the test set selection. This is calculated using bootstrap re-sampling for each optimizer run \cite{Clark:2011}. Variance due to optimizer instability was less than 0.1 in all cases. {\it p}-value of these tests are less than $0.05$ in all cases. Thus, comparison between two models is significant at 95\% CI if the difference in their score is more than the variance due to test set selection (Table 4).

\subsection{Analysis}
\label{sec:length}
\begin{figure}
    \centering
    \includegraphics[scale=0.30]{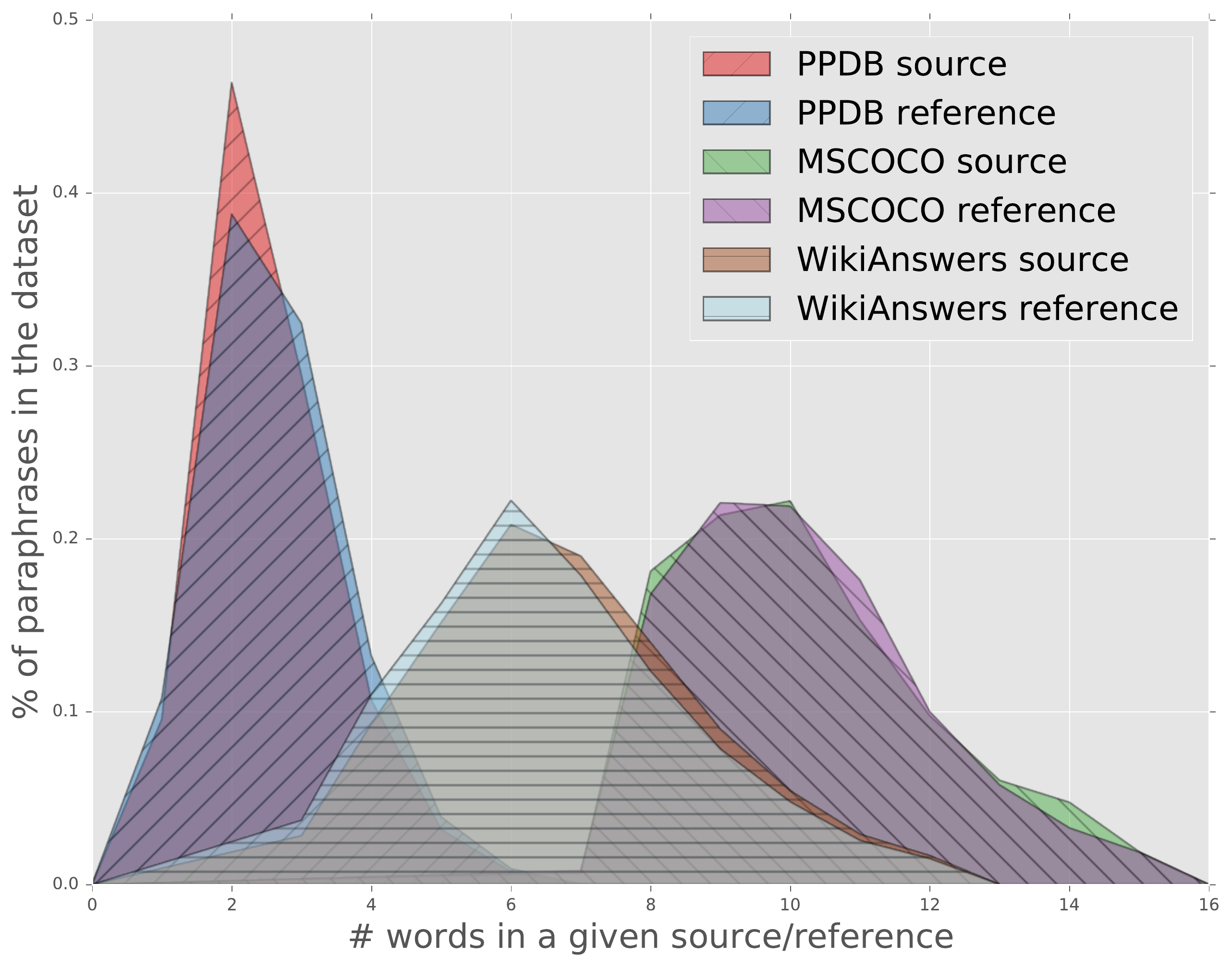}
       \caption{\label{fig:res1} Distribution of sequence length (in number of words) across datasets.}
		\vspace{3mm}
\end{figure}

    Scores on various metrics vary a lot across the datasets, which is understandable due to their inherent differences. \texttt{PPDB} contains very small phrases and thus does not score well with metrics like \texttt{BLEU} and \texttt{METEOR} which penalize shorter phrases. As shown in Figure~\ref{fig:res1}, more than $50\%$ of \texttt{PPDB} contains one or two words. This leads to a substantial difference between training and validation errors, as shown in Figure~\ref{fig:perplex}. The results demonstrate that deeper LSTMs consistently improve performance over shallow models. For beam size of 5 our model outperforms other models in all datasets. For beam size of 10, the attention-based model has a marginally better \emph{Emb Greedy} score than our model. When we look at the qualitative results, we notice that the bias in the dataset is exploited by the system which is a side effect of any form of learning on a limited dataset. We can see this effect in Table 5. For example, \emph{an OBJECT} is mostly paraphrased with \emph{an OBJECT} (e.g. \emph{bowl}, \emph{motorcycle}). Shorter sentences mostly generate shorter paraphrases and the same is true for longer sequences. Based on our results, the embedding-based metric correlates well with statistical metrics. Figure~\ref{fig:perplex} and the results from Table 5 suggest that perplexity is a good loss function for training paraphrase generation models. However, a more ideal metric to fully encode the fundamental objective of paraphrasing should also reward novelty and penalize redundancy during paraphrase generation, which is a notable limitation of the existing paraphrase evaluation metrics.
		

\global\long\def\ra#1{\global\long\def\arraystretch{#1}}
\begin{center}
 
\begin{table}
\small
    \centering \ra{1.1} %
    \begin{tabular}{@{}clrrrrcrrrr}
 
        \toprule
        &&& \multicolumn{3}{c}{Beam size = 5} & \phantom{abc} & \multicolumn{3}{c}{Beam size = 10}\tabularnewline
        \midrule
        \#Layers & \multicolumn{1}{c}{Model} & \scriptsize{\texttt{BLEU}}$\uparrow$ & \scriptsize{\texttt{METEOR}}$\uparrow$  & \scriptsize{Emb Greedy}$\uparrow$ & \scriptsize{\texttt{TER}}$\downarrow$  &  & \scriptsize{\texttt{BLEU}}$\uparrow$ & \scriptsize{\texttt{METEOR}}$\uparrow$  & \scriptsize{Emb Greedy}$\uparrow$ & \scriptsize{\texttt{TER}}$\downarrow$ \tabularnewline
        \toprule \multicolumn{11}{c}{\texttt{PPDB}}  \tabularnewline \midrule
 
        \multirow{2}{*}{2} & Sequence to Sequence & 12.5  & 21.3  & 32.55  & 82.9 & &  12.9 & 20.5 &  32.65 & 83.0 \tabularnewline
        & With Attention                          & 13.0  & 21.2  & 32.95  & 82.2 & &  13.8 & 20.6 &  32.29 & 81.9 \tabularnewline 
        \midrule
        \multirow{4}{*}{4} & Sequence to Sequence & 18.3  & \textbf{23.5}  & 33.18  & 82.7 & &  18.8 & \textbf{23.5} &  33.78 & 82.1 \tabularnewline
        & Bi-directional                           & 19.2  & 23.1  & 34.39  & 77.5 & &  19.7 & 23.2 &  34.56 & 84.4 \tabularnewline
        & With Attention                          & 19.9  & 23.2  & 34.71  & 83.8 & &  20.2 & 22.9 &  \textbf{34.90} & 77.1 \tabularnewline
        & \textbf{Residual LSTM}                   & \textbf{20.3}&  23.1& \textbf{34.77}  & \textbf{77.1}& & \textbf{21.2} & 23.0& 34.78 & \textbf{77.0} \tabularnewline
        \bottomrule
 
        \multicolumn{11}{c}{\texttt{WikiAnswers}}  \tabularnewline \midrule
 
        \multirow{2}{*}{2} & Sequence to Sequence & 19.2  & 26.1  &   62.65 & 35.1 & &  19.5 & 26.2 & 62.95 & 34.8 \tabularnewline
        & With Attention                          & 21.2  & 22.9  &   63.22 & 37.1 & &  21.2 & 23.0 & 63.50 & 37.0 \tabularnewline 
        \midrule
        \multirow{4}{*}{4} & Sequence to Sequence & 33.2  & 29.6  &   73.17 & 28.3 & & 33.5  & 29.6  & 73.19 & 28.3 \tabularnewline
        & Bi-directional                           & 34.0  & 30.8  &   73.80 & 27.3 & & 34.3  & 30.7  & 73.95 & 27.0 \tabularnewline
        & With Attention                          & 34.7  & 31.2  &   73.45 & 27.1 & & 34.9  & 31.2  & 73.50 & 27.1 \tabularnewline
        & \textbf{Residual LSTM}                  & \textbf{37.0}  & \textbf{32.2}  & \textbf{75.13}  & \textbf{27.0} & & \textbf{37.2}  & \textbf{32.2}  & \textbf{75.19} & \textbf{26.8} \tabularnewline
        \bottomrule
 
        \multicolumn{11}{c}{\texttt{MSCOCO}}  \tabularnewline \midrule
 
        \multirow{2}{*}{2} & Sequence to Sequence & 15.9  & 14.8  &   54.11 & 66.9 & &  16.5 & 15.4  & 55.81 & 67.1 \tabularnewline
        & With Attention                          & 17.5  & 16.6  &   58.92 & 63.9 & &  18.6 & 16.8  & 59.26 & 63.0 \tabularnewline 
        \midrule                                                                                           
        \multirow{4}{*}{4} & Sequence to Sequence & 28.2  & 23.0  &   67.22 & 56.7 & & 28.9  & 23.2  & 67.10 & 56.3 \tabularnewline
        & Bi-directional                           & 32.6  & 24.5  &   68.62 & 53.8 & & 32.8  & 24.9  & 68.91 & 53.7 \tabularnewline
        & With Attention                          & 33.1  & 25.4  &   69.10 & 54.3 & & 33.4  & 25.2  & \textbf{69.34} & 53.8 \tabularnewline
        & \textbf{Residual LSTM}                  & \textbf{36.7}  & \textbf{27.3}  &  \textbf{69.69} & \textbf{52.3} & & \textbf{37.0}  & \textbf{27.0}  & 69.21 & \textbf{51.6} \tabularnewline
        \bottomrule
 
    \end{tabular}
    \caption{Evaluation results on \texttt{PPDB}, \texttt{WikiAnswers}, and \texttt{MSCOCO} (Best results are in \textbf{bold}).}
\end{table}
\end{center}

\begin{table}
\small
\centering
\label{vartest}
\begin{tabular}{lllll}
\midrule
\multicolumn{1}{l}{Dataset} & \multicolumn{1}{c}{$\sigma^2$\texttt{[BLEU]}} & \multicolumn{1}{c}{$\sigma^2$\texttt{[METEOR]}} & $\sigma^2$\texttt{[TER]} & \multicolumn{1}{c}{$\sigma^2$[Emb Greedy]} \\
\midrule
\texttt{PPDB}                        & 2.8                      & 0.2                        & 0.4 & 0.000100 \\
\texttt{WikiAnswers}                     & 0.3                      & 0.1                        & 0.1 & 0.000017 \\
\texttt{MSCOCO}                      & 0.2                      & 0.1                        & 0.1 & 0.000013  \\
\midrule
\end{tabular}
\caption{Variance due to test set selection.}
\end{table}
 
\begin{center}
\begin{table}[htbp]
\scriptsize
\centering
\label{samples2}
\begin{tabular}{rlll}
\midrule
\multicolumn{1}{l}{}           & \multicolumn{1}{l}{{\texttt{PPDB}}}                                  & \multicolumn{1}{l}{{\texttt{WikiAnswers}}}                                                       & \multicolumn{1}{l}{{\texttt{MSCOCO}}}                              \\ \hline
\multicolumn{1}{r|}{\textbf{Source}}    & \multicolumn{1}{l||}{south eastern}    & \multicolumn{1}{l||}{what be the symbol of magnesium sulphate} & a small kitten is sitting in a bowl \\
\multicolumn{1}{r|}{\textbf{Reference}} & \multicolumn{1}{l||}{the eastern part} & \multicolumn{1}{l||}{chemical formulum for magnesium sulphate}  & a cat is curled up in a bowl        \\
\multicolumn{1}{r|}{\textbf{Generated}} & \multicolumn{1}{l||}{south east}       & \multicolumn{1}{l||}{do magnesium sulphate have a formulum}    & a cat that is sitting on a bowl     \\ \hline
\multicolumn{1}{r|}{\textbf{Source}}    & \multicolumn{1}{l||}{organized}        & \multicolumn{1}{l||}{what be the bigggest galaxy know to man} & an old couple at the beach during the day \\
\multicolumn{1}{r|}{\textbf{Reference}} & \multicolumn{1}{l||}{managed}          & \multicolumn{1}{l||}{how many galaxy be there in you known universe}  & two people sitting on dock looking at the ocean \\
\multicolumn{1}{r|}{\textbf{Generated}} & \multicolumn{1}{l||}{arranged}         & \multicolumn{1}{l||}{about how many galaxy do the universe contain}    & a couple standing on top of a sandy beach    \\ \hline
\multicolumn{1}{r|}{\textbf{Source}}    & \multicolumn{1}{l||}{counselling}      & \multicolumn{1}{l||}{what do the ph of acid range to} & a little baby is sitting on a huge motorcycle \\
\multicolumn{1}{r|}{\textbf{Reference}} & \multicolumn{1}{l||}{be kept informed} & \multicolumn{1}{l||}{a acid have ph range of what}  &  a little boy sitting alone on a motorcycle       \\
\multicolumn{1}{r|}{\textbf{Generated}} & \multicolumn{1}{l||}{consultations}    & \multicolumn{1}{l||}{how do acid affect ph}    &    a baby sitting on top of a motorcycle \\ \hline
\end{tabular}
\caption{Example paraphrases generated using the 4-layer Residual LSTM with beam size 5.} 
\end{table}
\end{center}

\section{Related Work}
Prior approaches to paraphrase generation have applied relatively different methodologies, typically using knowledge-driven approaches or statistical machine translation (SMT) principles. Knowledge-driven methods for paraphrase generation \cite{Madnani2010} utilize hand-crafted rules \cite{McKeown1983} or automatically learned complex paraphrase patterns \cite{Zhao2009}. Other paraphrase generation methods use thesaurus-based \cite{Hassan2007} or semantic analysis-driven natural language generation approaches \cite{Kozlowski2003} to generate paraphrases. In contrast, Quirk et al.,~\shortcite{quirk2004} show the effectiveness of SMT techniques for paraphrase generation given adequate monolingual parallel corpus extracted from comparable news articles. Wubben et al.,~\shortcite{Wubben2010} propose a phrase-based SMT framework for sentential paraphrase generation by using a large aligned monolingual corpus of news headlines. Zhao et al.,~\shortcite{Zhao08} propose a combination of multiple resources to learn phrase-based paraphrase tables and corresponding feature functions to devise a log-linear SMT model. Other models generate application-specific paraphrases \cite{Zhao2009}, leverage bilingual parallel corpora \cite{Bannard05} or apply a multi-pivot approach to output candidate paraphrases \cite{ZhaoWLL10}. 

    Applications of deep learning for paraphrase generation tasks have not been rigorously explored. We utilized several sources as potential large datasets. Recently, Weiting et al.~\shortcite{wieting2015ppdb} took the \texttt{PPDB} dataset (size $XL$) and annotated phrases based on their paraphrasability. This dataset is called \emph{Annotated-}\texttt{PPDB} and contains $3000$ pairs in total. They also introduced another dataset called \emph{ML-Paraphrase} for the purpose of evaluating bigram paraphrases. This dataset contains $327$ instances. Microsoft Research Paraphrase Corpus (MSRP) \cite{dolan2005microsoft} is another widely used dataset for paraphrase detection. MSRP contains $5800$ pairs of sentences (obtained from various news sources) accompanied with human annotations. These datasets are too small and therefore, we did not use them for training our deep learning models.

    To the best of our knowledge, this is the first work on using residual connections with recurrent neural networks. Very recently, we found that Toderici et al.~\shortcite{toderici2016full} used residual GRU to show an improvement in image compression rates for a given quality over JPEG.\ Another variant of residual network called \emph{DenseNet} \cite{huang2016densely}, which uses dense connections over every layer, has been shown to be effective for image recognition tasks achieving state-of-the-art results in CIFAR and SVHN datasets. Such works further validate the efficacy of adding residual connections for training deep networks.
    
\section{Conclusion and Future Work}
        In this paper, we described a novel technique to train stacked LSTM networks for paraphrase generation. This is an extension to sequence to sequence learning, which has been shown to be effective for various NLP tasks. Our model outperforms state-of-the-art models for sequence to sequence learning. We have shown that stacking of residual LSTM layers is useful for paraphrase generation, but it may not perform equally well for machine translation because not every word in a source sequence needs to be substituted for paraphrasing. Residual connections help retain important words in the generated paraphrases.
       
                                                We experimented on three different large scale datasets and reported results using various automatic evaluation metrics. We showed the use of the well-known \texttt{MSCOCO} dataset for paraphrase generation and demonstrated that the models can be trained effectively without leveraging the images. The presented experiments should set strong baselines for neural paraphrase generation on these datasets, enabling future researchers to easily compare and evaluate subsequent works in paraphrase generation.
        
                                                Recent advances in neural networks with regard to learnable memory \cite{sukhbaatar2015end,ntmalex} have enabled models to get one step closer to learning comprehension. It may be helpful to explore such networks for the paraphrase generation task. Also, it remains to be explored how unsupervised deep learning could be harnessed for paraphrase generation. It would be interesting to see if researchers working on image-captioning can employ neural paraphrase generation to augment their dataset.

\section*{Acknowledgment}
The authors would like to thank the anonymous reviewers for their valuable comments and feedback. The first author is especially grateful to Prof. James Storer, Brandeis University, for his guidance and Nick Moran, Brandeis University, for helpful discussions.
 
\bibliographystyle{acl}
\bibliography{ncp}
 

\end{document}